\documentclass[lettersize,journal]{IEEEtran}
\usepackage{amsmath,amsfonts}
\usepackage{algorithmic}
\usepackage{array}
\usepackage[caption=false,font=normalsize,labelfont=sf,textfont=sf]{subfig}
\usepackage{textcomp}
\usepackage{stfloats}
\usepackage{url}
\usepackage{color}
\usepackage{verbatim}
\usepackage{graphicx}
\usepackage[absolute,overlay]{textpos}
\def\BibTeX{{\rm B\kern-.05em{\sc i\kern-.025em b}\kern-.08em
		T\kern-.1667em\lower.7ex\hbox{E}\kern-.125emX}}
\usepackage{balance}
\usepackage{cite}
\usepackage{hyperref}
\usepackage{amsthm}  
\renewcommand{\baselinestretch}{1}
\newtheorem*{remark}{Remark}

\begin{document}
	
	\title{Residual Cross-Attention Transformer-Based Multi-User CSI Feedback with Deep Joint Source-Channel Coding}
	
	\author{Hengwei Zhang, Minghui Wu, Li Qiao, Ling Liu, Ziqi Han and Zhen Gao
	
	\thanks{The work was supported in part by the Natural Science Foundation of
		China (NSFC) under Grant 62471036, Beijing Natural Science Foundation
		under Grant L242011, Shandong Province Natural Science Foundation
		under Grant ZR2022YQ62, and Beijing Nova Program. (\textit{Corresponding
		author: Zhen Gao})
	}
	\thanks{Hengwei Zhang, Minghui Wu, Li Qiao and Ziqi Han are with the School of Information and Electronics, Beijing Institute of Technology, Beijing 100081, China (e-mail: zhanghengwei2001@163.com).
	}
	
	\thanks{Zhen Gao is with the State Key Laboratory of CNS/ATM and the MIIT Key
		Laboratory of Complex-Field Intelligent Sensing, Beijing Institute of Technology (BIT), Beijing 100081, China, also with BIT (Zhuhai), Zhuhai 519088, China, also with the Advanced Technology Research Institute, BIT (Jinan), Jinan 250307, China, and also with the Yangtze Delta Region Academy, BIT (Jiaxing), Jiaxing 314019, China (e-mail: gaozhen16@bit.edu.cn).}
		
	\thanks{ Ling Liu is with Guangzhou Institute of Technology, Xidian University, Guangzhou 510555, China (e-mail: liuling@xidian.edu.cn).}
		
	}
	


\maketitle

\begin{abstract}
	This letter proposes a deep-learning (DL)-based multi-user channel state information (CSI) feedback framework for massive multiple-input multiple-output systems, where the deep joint source-channel coding (DJSCC) is utilized to improve the CSI reconstruction accuracy. Specifically, we design a multi-user joint CSI feedback framework, whereby the CSI correlation of nearby users is utilized to reduce the feedback overhead. Under the framework, we propose a new residual cross-attention transformer architecture, which is deployed at the base station to further improve the CSI feedback performance. Moreover, to tackle the “cliff-effect” of conventional bit-level CSI feedback approaches, we integrated DJSCC into the multi-user CSI feedback, together with utilizing a two-stage training scheme to adapt to varying uplink noise levels. Experimental results demonstrate the superiority of our methods in CSI feedback performance, with low network complexity and better scalability.
\end{abstract}


\section{Introduction}
\IEEEPARstart{M}{assive} multiple-input multiple-output (MIMO) offers unprecedented spectral efficiency and network capacity in current wireless communication systems. At the base station (BS), to fully leverage the potential of massive MIMO, acquiring accurate downlink channel state information (CSI) is of vital importance. In frequency division duplex (FDD) massive MIMO systems, user equipment (UE) performs downlink channel estimation and subsequently feeds the CSI back to the BS\cite{guo2022overview}. However, the large number of antennas results in significant feedback overhead, posing a critical challenge for CSI feedback in massive MIMO systems.

To reduce the feedback overhead, CSI compression before transmission is essential \cite{guo2022overview}. The authors of \cite{rao2014distributed,gao2018compressive} employed compressed sensing (CS) algorithms in CSI feedback, but their reliance on channel sparsity often leads to suboptimal performance in practical scenario. The advent of deep learning (DL) has revolutionized CSI feedback \cite{wang2022transformer,cui2022transnet,wang2024crissnet,ma2021model,wang2024knowledge,guo2024deep,mashhadi2020distributed,xu2022deep,wen2018deep}, showing superior performance over conventional CS approaches.Recently, some studies in DL-based CSI feedback improved reconstruction performance by enhancing the network architecture, particularly through the utilization of attention mechanisms. Notable examples include reference \cite{wang2022transformer, cui2022transnet}, which employ the transformer \cite{vaswani2017attention} backbone for CSI feedback, and CRissNet \cite{wang2024crissnet}, which utilizes criss-cross attention for CSI feature extraction. 
Other studies conduct a DL-based joint channel estimation and CSI feedback \cite{wang2024knowledge,ma2021model}. Furthermore, the integration of deep joint source-channel coding (DJSCC) \cite{bourtsoulatze2019deep} into the task of CSI feedback \cite{xu2022deep} has enabled end-to-end optimization, enhancing performance in practical uplink transmission.

To further improve the CSI reconstruction performance under limited feedback overhead, some researchers considered to explore the correlation in downlink CSI among nearby UEs served by the same BS.  This correlation arises from shared physical propagation environments, as illustrated in \cite{rao2014distributed}, demonstrating that CSI matrices from different UEs exhibit very similar support sets in the angular domain. Nevertheless, the distributed nature of UEs makes joint compression unfeasible, requiring joint processing at the BS. The authors of\cite{guo2024deep} proposed to incorporate an additional decoder at the BS for reconstructing the magnitude of CSI from nearby UEs, while the phase are compressed and recovered independently, and the additional decoder also increase complexity. The authors of \cite{mashhadi2020distributed} proposed adding summation-based fusion branches between decoders of nearby UEs to fuse information from other UEs, which still left room for effectively extracting and leveraging CSI correlation. Moreover, as the number of UEs increases, both the structural and computational complexity issues of the network become more critical. In the field of multimodal learning \cite{xu2023multimodal}, cross-attention has shown promise in multimodal interaction with effectiveness and lower computational complexity. However, its application in multi-user CSI feedback scenarios remains unexplored. Furthermore, without considering DJSCC \cite{xu2022deep}, the authors of \cite{rao2014distributed,guo2024deep,mashhadi2020distributed} all use separate source-channel coding (SSCC) for uplink CSI transmission, which may suffer from “cliff-effect” in the practical scenario. 

In this letter, we propose a novel residual cross-attention multi-user network (RCA-MUNet) - a new framework for multi-user CSI feedback. Our contributions are as follows:

\begin{itemize}
	\item A well-deigned RCA-Block is proposed to effectively extract and fuse features from CSI of nearby UEs, which is incorporated to the transformer-based backbone at the BS to enhance CSI reconstruction. Unlike existing methods, the RCA-MUNet can serve  any number of UEs (at least two UEs) without introducing additional complexity or changing architecture.
	\item On the basis of RCA-MUNet, we further integrate the emerging DJSCC to multi-user CSI feedback. We also utilize a new network training scheme to adapt to varying uplink signal-to-noise ratios (SNR).
	\item Extensive simulations are conducted under different compression ratios (CR), uplink SNRs, and numbers of UEs, demonstrating the superiority of the proposed method.
\end{itemize}   
\textit{Notations:} Boldface lower and upper-case symbols denote column vectors and matrices, respectively. Superscripts ${\left(  \cdot  \right)^{\rm{T}}}$ and ${\left(  \cdot  \right)^{\rm{H}}}$ denote the transpose and conjugate transpose operators, respectively. $\Vert \cdot \Vert_F$ represents the Frobenius norm of a matrix.
\vspace{-20pt}

\section{Channel Model}
\begin{figure}[t]
	\centering
	\includegraphics[width=0.35\textwidth]{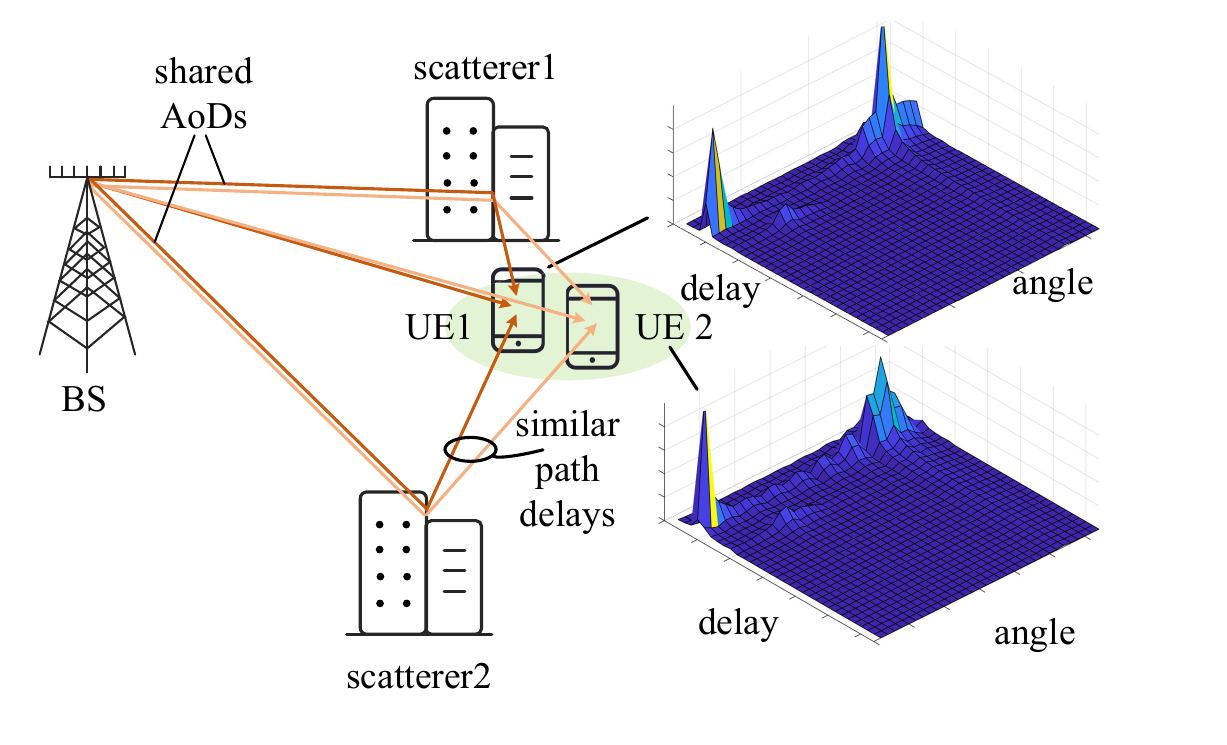}
	\vspace{-10pt}  
	\caption{Illustration of CSI correlation from nearby UEs with shared scatterers.}
	\label{fig:CSI Correlation}
	\vspace{-20pt}  
\end{figure}

We consider a typical FDD massive MIMO-OFDM system. The BS is equipped with a uniform linear array (ULA) with ${N_t}$ antennas, serving $M$ single-antenna UEs, as shown in Fig. \ref{fig:CSI Correlation}. Orthogonal frequency division modulation (OFDM) with ${N_c}$ subcarriers is adopted.
The downlink channel between the BS and the $m$th UE on the $k$th subcarrier can be denoted as ${\bf{h}}_{k,m} \in {\mathbb{C}^{{N_t} \times {N_r}}}$, where $m \in \left\{ {1, \ldots ,M} \right\}$, $k \in \left\{ {1, \ldots ,{N_c}} \right\}$. According to the classic multipath channel model applied in \cite{mashhadi2020distributed}, the downlink channel ${\bf{h}}_{k,m}$ can be expressed as
\begin{equation}
	\label{Eq:MIMOChannel}
	{\bf{h}}_{k,m} = \sqrt {\frac{{{N_t}}}{{{L_m}}}} \sum\limits_{l = 1}^{{L_m}} {\alpha _{l,m}{e^{ - j2\pi \frac{k}{{{N_c}}}\tau _{l,m}{f_s}}}} {{\bf{a}}_t}\left( {\phi _{l,m}} \right),
\end{equation}
where $L$ denotes the number of multi-path components (MPCs) for the 
$m$th UE, ${\alpha _{l,m}}$ and ${\tau _{l,m}}$ represent the propagation gain and path delay of the $l$-th MPC, and ${f_s}$ is the system bandwidth. Given that the antenna spacing is half-wavelength, the transmit steering vector ${{\bf{a}}_t}\left( \phi  \right) = \left[ {1,e^{-j\pi \sin \left ( \phi  \right ) } , \cdots ,e^{-j\pi \left ( N_{t}-1  \right ) \sin \left ( \phi  \right ) } } \right]^{\rm{T}}$, where ${\phi}$ is the angle of departure (AoD). The spatial-frequency domain CSI of each UE ${{\bf{\tilde H}}_m} = {\left[ {{\bf{h}}_{1,m},{\bf{h}}_{2,m}, \ldots ,{\bf{h}}_{{N_c},m}} \right]^{\rm{T}}} \in \mathbb{C}{^{{N_c} \times {N_t}}}$. The angle-delay domain CSI ${{\bf{H}}_m} = {{\bf{F}}_d}{{\bf{\tilde H}}_m}{\bf{F}}_a^{\rm{H}}$ exhibits obvious sparsity, where ${{{\bf{F}}_d} \in {\mathbb{C}^{{N_c} \times {N_c}}}}$and ${{{\bf{F}}_a} \in {\mathbb{C}^{{N_t} \times {N_t}}}}$are DFT matrices. For spatially neighboring UEs, the shared AoDs of UEs lead to very similar angular support sets. Furthermore, the spatial consistency indicates that nearby UEs also have very similar path delays \cite{3gpp2017study}. Therefore, angle-delay domain CSI matrices of nearby UEs exhibit significant correlation. Fig. \ref{fig:CSI Correlation} shows two CSI magnitude samples of two nearby UEs, which have similar support sets indices and different amplitude.  

\section{Proposed Multi-User CSI Feedback Scheme}
We first introduce the overall architecture of the proposed transformer-based DJSCC multi-user CSI feedback framework, and then provide details of the residual cross-attention multi-user CSI joint processing module.
\vspace{-10pt}
\subsection{Proposed Multi-User CSI Feedback Network Architecture}
As shown in Fig.\ref{fig:feedback framework}, in the multi-user scenario, UEs first perform CSI compression independently and feed the compressed CSI back to the BS through the uplink control link for reconstruction. In practice, since the MPCs are concentrated in the front limited delay region, only the first ${N_a}$ rows are truncated for CSI feedback \cite{guo2022overview}, defined as ${{{\bf{H}}_{a,m}} \in {\mathbb{C}^{{{N}_a} \times {N_t}}}}$. The truncated angular-delay domain CSI ${{\bf{H}}_{a,m}}$ is divided into ${N_t}$ vectors, and we concatenate the real and imaginary parts of the vectors to form the input matrix. We then convert the input matrix to embeddings ${{\bf{X}}_m} \in {\mathbb{R}^{{N_t} \times {d}}}$ through linear projection, where ${d}$ is the transformer model dimension. Positional encodings are then injected into the embeddings \cite{vaswani2017attention}. Subsequently, the embeddings pass through the backbone, which is composed of ${L_1}$ transformer layers, and are then projected back to size ${N_t} \times 2{N_c}$ in the last layer. The output embeddings are vectorized into a vector of size $2{N_t}{N_c}$ and then compressed to ${{\bf{v}}_m} \in {\mathbb{R}^{k}}$ by a fully connected layer.
We then adopt DJSCC for CSI feedback to enhance the end-to-end reconstruction quality, in which the aforementioned encoder network is treated as the joint source-channel encoder. The compressed real-valued vector ${{\bf{v}}_m}$ is mapped to a complex-valued vector ${{\bf{s}}_m} \in {\mathbb{C}^{k/2}}$ and then power-normalized. Each element of ${{\bf{s}}_m}$ is assigned to a subcarrier as a constellation symbol and transmitted orthogonally through the uplink control link. Without loss of generality, we consider the link as an additive white Gaussian noise (AWGN) channel. The received symbols at the BS is ${{\bf{\bar s}}_m} = {{\bf{s}}_m} + {{\bf{z}}_m}\in {\mathbb{C}^{k/2}}$, where ${{\bf{z}}_m} \sim {\cal C}{\cal N}({\bf{0}},{N_{0,m}}{\bf{I}})$, a complex Gaussian distribution with mean $\bf{0} $ and covariance matrix ${N_{0,m}}{\bf{I}}$. ${\bf{I}}\in {{\mathbb{C}}^{k/2 \times k/2}} $ is the identity matrix. The received symbol ${{\bf{\bar s}}_m}$ from each UE is mapped into a real-valued vector ${{\bf{\bar v}}_m}$ for decompression. 

At the BS, embedding and positional encoding are also added to the decoder to maintain structural symmetry, aligning with \cite{wang2022transformer}. Subsequently, a backbone with ${L_2}$ transformer layers and ${L_3}$ residual cross-attention blocks (RCA-Block) is applied, which is treated as the joint source-channel decoder. The real-valued vector ${\bf{\bar v}}_m$ is projected to size $2{N_t}{N_c}$ by a fully connected layer and then reshaped to size ${N_t} \times 2{N_c}$. After embedding and positional encoding, the CSI from each UE is first independently reconstructed through their individual transformer layers. The outputs are then shared and processed jointly by RCA-Blocks to get the enhanced CSI ${{{\bf{\hat H}}}_m}$. 

The encoders and decoders of all the UEs, together with the uplink control link, are all taken account into network training. The entire process is optimized in an end-to-end manner. We take the original CSI of $M$ UEs as input, and then output the reconstructed CSI of all $M$ UEs. The loss function for network training ${\rm{MS}}{{\rm{E}}_{{\rm{Loss}}}} = \sum\limits_{m = 1}^M {\left\| { {{{\bf{H}}_{a,m}} - {{{\bf{\hat H}}}_m}}} \right\|_F^2}$, which is the sum of the reconstruction mean square error (MSE) of all the UEs.

Furthermore, to adapt to various uplink SNRs, we adopt a two-stage training strategy. The multi-user network is first pretrained across a broad range of SNRs, and then fine-tuned at each specific SNR. Note that we also adopt a parameter sharing scheme, in which the encoders and decoders of UEs share the same parameters, thus the complexity of the multi-user network remains unchange as the number of UE increases.
\vspace{-20pt}
\subsection{Residual Cross-Attention Multi-User CSI Joint Processing}
\begin{figure*}[t]
	\centering
	\includegraphics[width=0.75\textwidth]{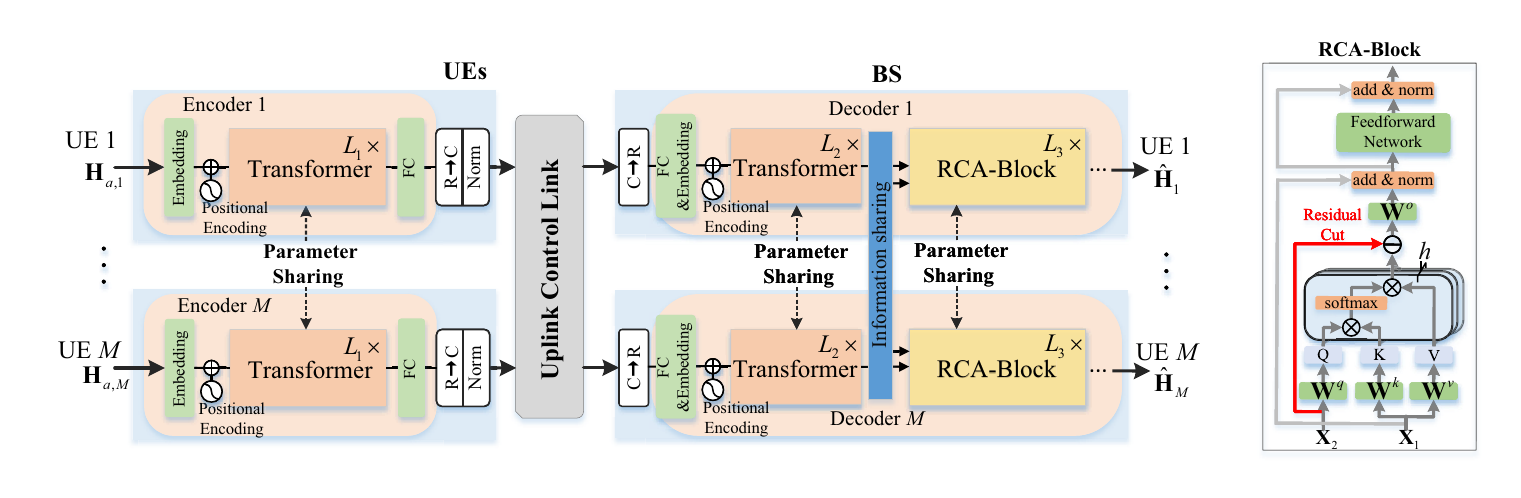}
	\vspace{-15pt}
	\caption{Proposed Transformer-based DJSCC multi-user CSI feedback framework and architecture of RCA-Block.}
	\vspace{-15pt}  
	\label{fig:feedback framework}
\end{figure*}
Due to the typically distributed nature of UEs, joint processing is only available at the BS. To exploit the CSI correlation of nearby UEs, we focus on leveraging the emerging cross attention mechanism. Cross-attention has been extensively utilized in multi-modal learning to extract and exploit features from correlated modalities\cite{xu2023multimodal}. This proven effectiveness motivates us to employ cross-attention for multi-user CSI feedback. Typical cross-attention can be expressed as \cite{vaswani2017attention}
\begin{equation}
	\label{Eq:dot-product attention}
	{\rm{Attention}}\left( {\bf{Q},K,V} \right) = {\rm{softmax}}\left( {\frac{{{\bf{Q}}{{\bf{K}}^{\rm{T}}}}}{{\sqrt {d} }}} \right)\bf{V},
\end{equation}
where the query $\bf{Q}$ is derived from one modality, key $\bf{K}$ and value $\bf{V}$ are obtained from another modality. $\bf{Q}$, $\bf{K}$ and $\bf{V}$ are all real-valued tensor of dimension ${N_t} \times {d}$. The dot-product attention calculates attention weights by matching $\bf{Q}$ with $\bf{K}$ and score its similarity, assigns higher attention to $\bf{V}$ with greater correlation, extracting common information between the two modalities. However, different from the typical cross-attention, our goal is to extract \textit{complementary information} between UEs rather than common information. To achieve this, we propose a new residual cross-attention architecture.

We first consider the case of two UEs. Let ${\bf{X}}_1$ represent the embedding of the current UE, and let ${\bf{X}}_2$ denote the embedding of the other UE.  ${\bf{X}}_1$ and ${\bf{X}}_2$ are all real-valued tensor of dimension ${N_t} \times {d}$. ${\bf{X}}_2$ is first passed through a linear projection to obtain ${\bf{Q}} \in {\mathbb{R}^{{N_t} \times {d}}}$, while ${\bf{K}} \in {\mathbb{R}^{{N_t} \times {d}}}$ and ${\bf{V}} \in {\mathbb{R}^{{N_t} \times {d}}}$ are derived from $\bf{X_1}$, expressed as
\begin{equation}
	\label{Eq:qkv projection}
	{\bf{Q}} = {{\bf{X}}_2}{{\bf{W}}^q},{\bf{K}} = {{\bf{X}}_1}{{\bf{W}}^k},{\bf{V}} = {{\bf{X}}_1}{{\bf{W}}^v},
\end{equation}
${{\bf{W}}^q},{{\bf{W}}^k},{{\bf{W}}^v}$ are all real-valued matrix of size $d  \times d $,  the attention map is calculated as \eqref{Eq:dot-product attention}. In contrast to the basic cross-attention, a residual cut is added as follows:
\begin{equation}
	\label{Eq:residual cut}
	{\bf{\tilde X}} = {{\bf{X}}_2} - {\rm{Attention}}({\bf{Q}},{\bf{K}},{\bf{V}}),
\end{equation}
where this operation suppresses the correlated features in ${{\bf{X}}_2}$, extracting the complementary information between the two UEs. ${\bf{\tilde X}}$ is then processed by a linear projection ${{\bf{W}}^o} \in {\mathbb{R}^{{d} \times {d}}} $ and get the output of residual cross-attention ${{\bf{X}}_{R}}$, namely
\begin{equation}
	\label{Eq:output projection}
	{{\bf{X}}_R} = {{\bf{\tilde X}}}{{\bf{W}}^o}.
\end{equation}

Based on the proposed residual cross-attention, we design RCA-Block to extract and fuse complementary information from multi-user CSIs, as shown in Fig.\ref{fig:feedback framework}. Specifically, to enhance the model's capacity, we employ the multi-head attention mechanism. ${\bf{Q}}$, ${\bf{K}}$, and ${\bf{V}}$ are reshaped into $h$ heads, denoted as ${\bf{Q}}_i$, ${\bf{K}}_i$, ${\bf{V}}_i\left( {1 \le i \le h} \right)$, which are of shape $N_t \times {d}/h $. Each attention head can capture the correlations of nearby users' CSI matrices in different subspaces, thereby enhancing the network's capacity to extract complementary information. Similarly, the attention output of each head ${\bf{M}}_i$ is calculated in parallel as \eqref{Eq:dot-product attention} and then concatenated as
\begin{equation}
	\label{Eq:mhca}
	{\bf{\tilde X}}^h = {\rm{concat}}\left( {\bf{M}}_1, \ldots ,{\bf{M}}_h \right),
\end{equation}
where ${\bf{\tilde X}}^h \in {R^{{N_t} \times {d}}}$. We then compute the output of multi-head residual cross-attention ${\bf{X}}_R^h$ through \eqref{Eq:residual cut} and \eqref{Eq:output projection}, namely
\begin{equation}
	\label{Eq:mh output projection}
	{\bf{X}}_R^h = \left( {{{\bf{X}}_2} - {{{\bf{\tilde X}}}^h}} \right){{\bf{W}}^o}.
\end{equation}

To improve the convergence performance, a skip connection and layer normalization (LayerNorm) is added. Additionally, a feedforward network (FFN) with a skip connection \cite{vaswani2017attention} is added after the multi-head residual cross-attention to produce the final output ${\bf{Z}}_o$ of the module:
\begin{align}
	{\bf{Z}} &= {\rm{LayerNorm}}\left( {{\bf{X}}_R^h + {{\bf{X}}_1}} \right), \label{Eq:skip connection} \\
	{{\bf{Z}}_o} &= {\rm{LayerNorm}}\left( {{\bf{Z}} + {\rm{FFN}}\left( {\bf{Z}} \right)} \right). \label{Eq:FFN}
\end{align}

For more than two UEs, we compute the element-wise sum of the embeddings of other UEs except UE $i$ and take the average, namely ${{\bf{X}}_2} = \left( {\sum\nolimits_{m = 1,m \ne i}^M {{{\bf{X}}_m}} } \right)/\left( {m - 1} \right)$, to aggregate information from other UEs. ${\bf{X}}_1$ and ${\bf{X}}_2$ is then fed into the aforementioned module for joint processing. 

The RCA-Block adopts a transformer-based structure, which can be integrated into the multi-user network by replacing the original transformer layers. Unlike existing methods, our method can adapt to any number of UEs without changing architecture, thereby enhancing deployment flexibility. This will be verified in Section \ref{sec:simulations} through simulations.

\begin{remark}
	We represent the CSI information of two nearby UE by sets ${I_1}$ and ${I_2}$, with their intersection ${I_{1\cap 2}} $. The independently recovered CSI information sets are  ${I_1^{\prime}}$ and ${I_2^{\prime}}$. Due to the independence of compression and feedback, the information retained in ${I_1^{\prime}}$ and ${I_2^{\prime}}$ from ${I_{1\cap2}}$, denote as ${C_1 = I_1^{\prime} \cap {I_{1\cap2}}}$ and ${C_2 = I_2^{\prime} \cap {I_{1\cap2}}}$ , is also different. Thus the union of $C_1$ and $C_2$ should be greater than or equal to either A or B, while being less than or equal to ${I_{1\cap2}}$, i.e.,
	\begin{equation}
		\label{Eq:multiuser information}
		\max \left( {C_1, C_2} \right) \le C_1 \cup C_2 \le {I_{1\cap2}}.
	\end{equation}
	
	In other words, one UE can retain the information loss of another UE caused by compression and noise, and vice versa. The key point lies in how to extract and exploit complementary information from other UEs that is useful for the current UE.
\end{remark}
\section{Simulation Results}
\label{sec:simulations}
In this section, we compare the proposed method with state-of-the-art methods, analyze its performance under various conditions and conduct ablation study to evaluate the effectiveness of the proposed method. Finally, we examine the superiority of the proposed DJSCC compared to conventional SSCC.

The downlink CSI dataset is generated by QuaDRiGa \cite{jaeckel2014quadriga} in accordance with the 3GPP TR 38.901 \cite{3gpp2017study}. Specifically, the urban macrocell (Uma) scenario with a downlink center frequency 6.7 GHz is considered. The number of subcarriers is $N_c = 1024$, which is then truncated to ${\bar N_c}=32$. The antenna configuration with $N_t=32$ transmit antennas at the BS and $N_r=1$ receive antenna at the UEs is adopted. We generate a spatially correlated multi-user CSI dataset for 4 UEs within a 300m $\times$ 300m cell, with a inter-user distance constraint of less than 10m. The training, validation and testing datasets consists of 80,000, 20,000, and 10,000 samples, respectively. Each CSI sample is power-normalized. The network is trained using the Adam optimizer for 2000 epochs, with the learning rate scheduled by the NoamOpt \cite{vaswani2017attention} method. We adopted the average reconstruction normalized mean square error (NMSE) of each UE, namely ${\rm{NMSE}} = \left( {\sum\limits_{m = 1}^M {\left\| {{{{\bf{H}}_{a,m}} - {{{\bf{\hat H}}}_m}}} \right\|_2^2/\left\| {{{{\bf{H}}_{a,m}}} } \right\|_2^2} } \right)/M$, as the CSI feedback performance metric.
\vspace{-10pt}
\subsection{Comparison with State-Of-The-Art Methods}
\renewcommand{\arraystretch}{0.9}
\begin{table*}[!t]
	\captionsetup{labelfont={color=red}}
	\renewcommand{\thetable}{I}
	{
		\centering
		\caption{{CSI Feedback Performance NMSE Comparison at Different CRs and SNRs (2 UEs are considered except for \textbf{BL1} and \textbf{BL2})}}
		\setlength{\tabcolsep}{8.8pt} 
		\begin{tabular}{c|c|c|c|c|c|c|c|c|c|c|c|c}  
			\hline
			CR & \multicolumn{3}{c|}{1/4} & \multicolumn{3}{c|}{1/8} & \multicolumn{3}{c}{1/16} & \multicolumn{3}{c}{1/32} \\ \hline
			
			SNR & 20 dB & 10 dB & 0 dB & 20 dB & 10 dB & 0 dB & 20 dB & 10 dB & 0 dB & 20 dB & 10 dB & 0 dB \\ \hline
			
			\textbf{BL1} & -24.58 & -20.55 & -13.81 & -20.26 & -16.32 & -11.01 & -16.76 & -14.69 & -7.47 & -13.59 & -11.03 & -4.21 \\ 
			
			\textbf{BL2} \cite{xu2022deep} & -19.17 & -17.45 & -11.41 & -16.51 & -14.99 & -8.97 & -14.37 & -12.70 & -6.16 & -10.84 & -9.36 & -3.36 \\ 
			
			\textbf{BL3} \cite{mashhadi2020distributed}& -25.26 & -20.88 & -14.14 & -21.99 & -17.89 & -11.82 & -18.08 & -15.69 & -8.75 & -14.77 & -12.44 & -5.41 \\ 
			
			\textbf{BL4} & -25.88 & -20.74 & -13.89 & -21.00 & -17.41 & -11.40 & -17.20 & -14.94 & -8.38 & -14.43 & -11.64 & -5.28 \\
			
			\textbf{BL5} & -24.68 & -20.56 & -13.91 & -20.25 & -16.49 & -11.14 & -16.95 & -15.00 & -7.68 & -13.65 & -11.01& -4.22 \\
			
			\textbf{Ours} & \textbf{-26.91} & \textbf{-21.60} & \textbf{-14.76} & \textbf{-22.97} & \textbf{-18.74} & \textbf{-12.24} & \textbf{-19.24} & \textbf{-16.76} & \textbf{-8.85} & \textbf{-16.04} & \textbf{-13.01} & \textbf{-5.61} \\ \hline
			
	\end{tabular}
	\vspace{-10pt}  
	\label{tb:results_all}
	}
\end{table*}

We first compare the CSI reconstruction performance of the proposed RCA-MUNet (\textbf{Ours}) with that of the following baselines. In \textbf{Ours}, we set the layers of network as $L_1 = 4$ and $L_2 = L_3 = 3$. \textbf{BL1} simplifies the RCA-MUNet to single-user network, which employs the encoder of \textbf{Ours} and replaces the RCA-Blocks in \textbf{Ours} with transformer layers, ensuring the same depth. \textbf{BL2} is the DJSCC-based single-user CSI feedback proposed in \cite{xu2022deep}, named ADJSCC-CSINet+. \textbf{BL3} extends \textbf{BL1} to a multi-user case by incorporating the joint reconstruction method in \cite{mashhadi2020distributed}, where fusion branches are added to the last $L_3 = 3$ layers of the decoder. \textbf{BL4} conducts an ablation study by replacing the residual cross-attention in the RCA-Blocks of \textbf{Ours} with typical cross-attention \cite{vaswani2017attention}. \textbf{BL5} replicates the network architecture of \textbf{Ours} while introducing a key difference: at the BS, each UE utilizes their own embeddings to generate query $\bf{Q}$ for cross-attention without interacting with other UEs. \textbf{CDL-OP} is the dictionary learning based CSI feedback method in \cite{CDLOP}. We consider a noise-free feedback for \textbf{CDL-OP} align with \cite{CDLOP}.

In TABLE \ref{tb:results_all}, we evaluate the CSI reconstruction NMSE (dB) of different methods at SNRs (20 dB, 10 dB, 0 dB) and CRs (1/4, 1/8, 1/16, 1/32), the multi-user methods \textbf{BL3}, \textbf{BL4}, \textbf{BL5} and \textbf{Ours} are tested under a 2-UE scenario. The NMSE of \textbf{CDL-OP} with perfect CSI feedback are -11.02dB, -7.59dB, -4.71dB, -1.72dB, respectively. It is evident that DL-based schemes \textbf{BL1}-\textbf{BL5} and \textbf{Ours} with uplink channel noise significantly outperform the noise-free \textbf{CDL-OP}, which reveals the superiority of DL-based CSI feedback over conventional CS-based approaches, aligning with the analysis in \cite{wen2018deep}. The results also show that the \textbf{Ours} method achieves significant gains over \textbf{BL1} across all CRs and SNRs, demonstrating the effectiveness of the proposed multi-user joint processing mechanism. We can also observe that \textbf{Ours} outperforms \textbf{BL3} in terms of recovery quality. Moreover, in all cases, the NMSE of \textbf{BL1} outperforms \textbf{BL2}, further illustrates the superiority of the proposed residual cross-attention transformer-based DJSCC framework.

Regarding the neural network complexity of the multi-user methods, in TABLE \ref{tb:complexity}, we calculate the floating-point operations (FLOPs) and parameters of \textbf{BL1}, \textbf{BL2}, \textbf{BL3}, and \textbf{Ours} (\textbf{BL4}, \textbf{BL5} has nearly the same complexity with \textbf{Ours}). “M” represents million. Comparing with the transformer-based \textbf{BL1}, the CNN-based \textbf{BL2} reduces parameters and FLOPs to 57$\%$ and 22$\%$, respectively. However, both architectures maintain comparable complexity levels, with \textbf{BL1} achieving significantly better NMSE performance than \textbf{BL2}. Comparing with the single-user network \textbf{BL1}, the params of \textbf{Ours} do not increase, while \textbf{BL3} exhibits evident parameter growth. This is because the RCA-Block in \textbf{Ours} replaces the original layers without introducing additional parameters, whereas \textbf{BL3} requires $M(M-1)$ additional branches for $M$ users. Furthermore, the FLOPs of \textbf{Ours} increases linearly with the number of UEs, while the FLOPs of \textbf{BL3} grows faster than linear due to the calculations in additional branches. The results show the superiority of \textbf{Ours} in terms of complexity. 

In TABLE I, we also conducted ablation study \textbf{BL4}, \textbf{BL5} to further validate the effectiveness of the proposed residual cross-attention. We first compare \textbf{BL1}, \textbf{BL4}, and \textbf{Ours}. Both \textbf{BL4} and \textbf{Ours} extract information from other UEs. \textbf{BL4} enhances performance by extracting and enhancing inter-user common features from nearby UEs via cross-attention. However, \textbf{BL4} cannot compensate for information loss like the residual cross-attention in \textbf{Ours}, resulting in inferior performance. Subsequently, we compare \textbf{BL1}, \textbf{BL5}, and \textbf{Ours}. \textbf{BL1} and \textbf{BL5} exhibit nearly identical performance,  with \textbf{BL5} slightly outperforming \textbf{BL1} by 0.1–0.2 dB. This is because the residual cut in \textbf{BL5} plays a role in feature reallocation, enabling the network to focus on less-attended features in self-attention. Nevertheless, \textbf{BL5} fails to recover the information loss, yielding lower performance than \textbf{Ours}. The results highlight the contribution of the residual cut in \textbf{Ours}.
\vspace{-10pt}
\begin{table}[t]
	\centering
	{
		\setlength{\tabcolsep}{2.3pt}
		\caption{Neural Network Complexity Comparison}
		\begin{tabular}{c|c|c|c|c|c}
			\hline
			~ & \textbf{BL1} & \textbf{BL2} & \textbf{BL3} (2 UE) & \textbf{Ours} (2 UE) & \textbf{Ours} (4 UE) \\ \hline
			FLOPs & 144.96 M & 82.47 M & 346.60 M & 289.83 M & 579.66 M \\ \hline
			parameters & 5.55 M & 1.24 M & 6.44 M & 5.55 M & 5.55 M \\ \hline
		\end{tabular}
	
	\label{tb:complexity}
	}
\end{table}

\subsection{Performance of Proposed Schemes Across Conditions}
\begin{figure}[t]
	{
	\centering
	\vspace{-10pt}
	\includegraphics[width=0.4\textwidth]{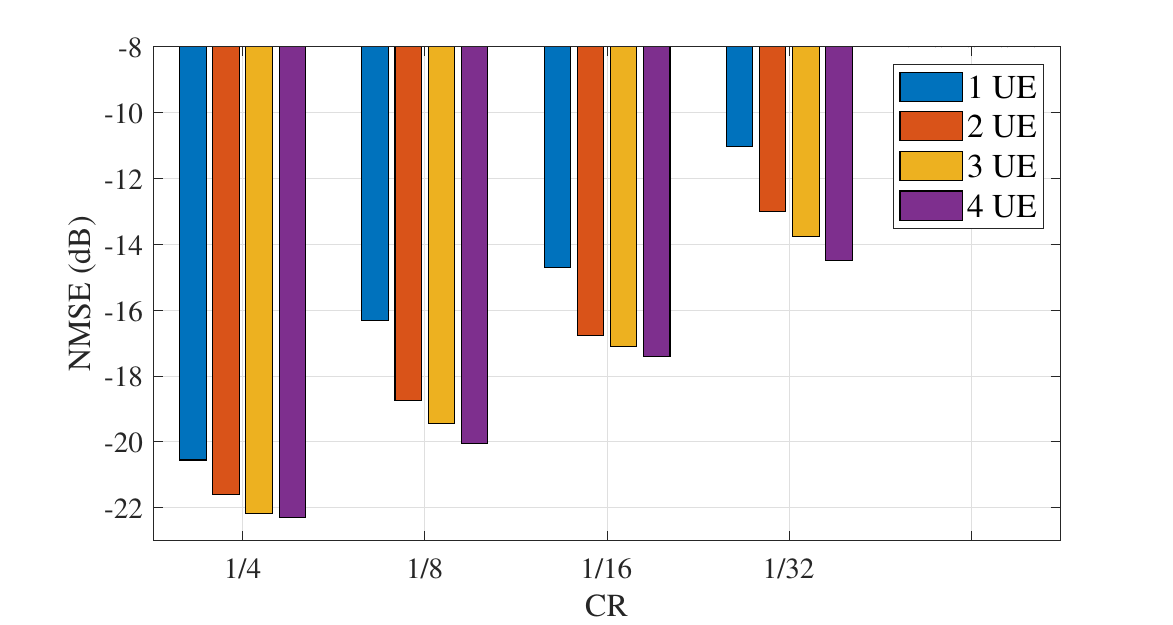}
	\vspace{-10pt}  
	\caption{NMSE performance for single-user network \textbf{BL1} and multi-user framework \textbf{Ours} with different numbers of UEs.}
	\vspace{-20pt}  
	\label{fig:34 result}
	}
\end{figure}
In TABLE \ref{tb:results_all}, we further analyze the impact of CR and SNR on multi-user processing gain by comparing \textbf{Ours} with \textbf{BL1}. At high CRs, CSI reconstruction is near optimal, with minimal loss and limited multi-user gains. As CR decreases, the growing compression loss allows for greater compensation from other UEs' complementary information. However, at very low CRs, the severe information loss of each UE reduces the available complementary information, leading to the decline of performance gain. Furthermore, since both compression and feedback inherently lead to information loss, the trend of gains under different SNRs can be explained in the same way. As the SNR decreases, the available complementary information between users diminishes, leading to reduced performance gains. This degradation becomes particularly pronounced at low SNR (e.g., 0 dB), where both UEs experience severe information loss, resulting in a significant decline in performance gains.

Additionally, in Fig. \ref{fig:34 result}, we extend the numbers of UE of \textbf{Ours} to 3 and 4 and make comparison with \textbf{BL1}. Since the trends observed across different SNR levels are consistent, we conduct the tests at SNR $=10$ dB. From Fig. \ref{fig:34 result}, it can be observed that as the number of UEs increases, the NMSE improves at each CR, demonstrating that the network can scale to more than two UEs and further enhance reconstruction performance. However, the gain improves slower as the number of UE increases. This is due to the marginal effect of multi-user joint processing as the number of UEs grows. Unlike \textbf{BL3}, which requires adding varying numbers of branches for different numbers of UEs, \textbf{Ours} uses the same network for varying numbers of UE, thus offering better scalability.
\vspace{-12pt}
\begin{figure}[t]
	\centering
	\begin{minipage}[b]{0.45\columnwidth}
		\centering
		\includegraphics[width=\textwidth]{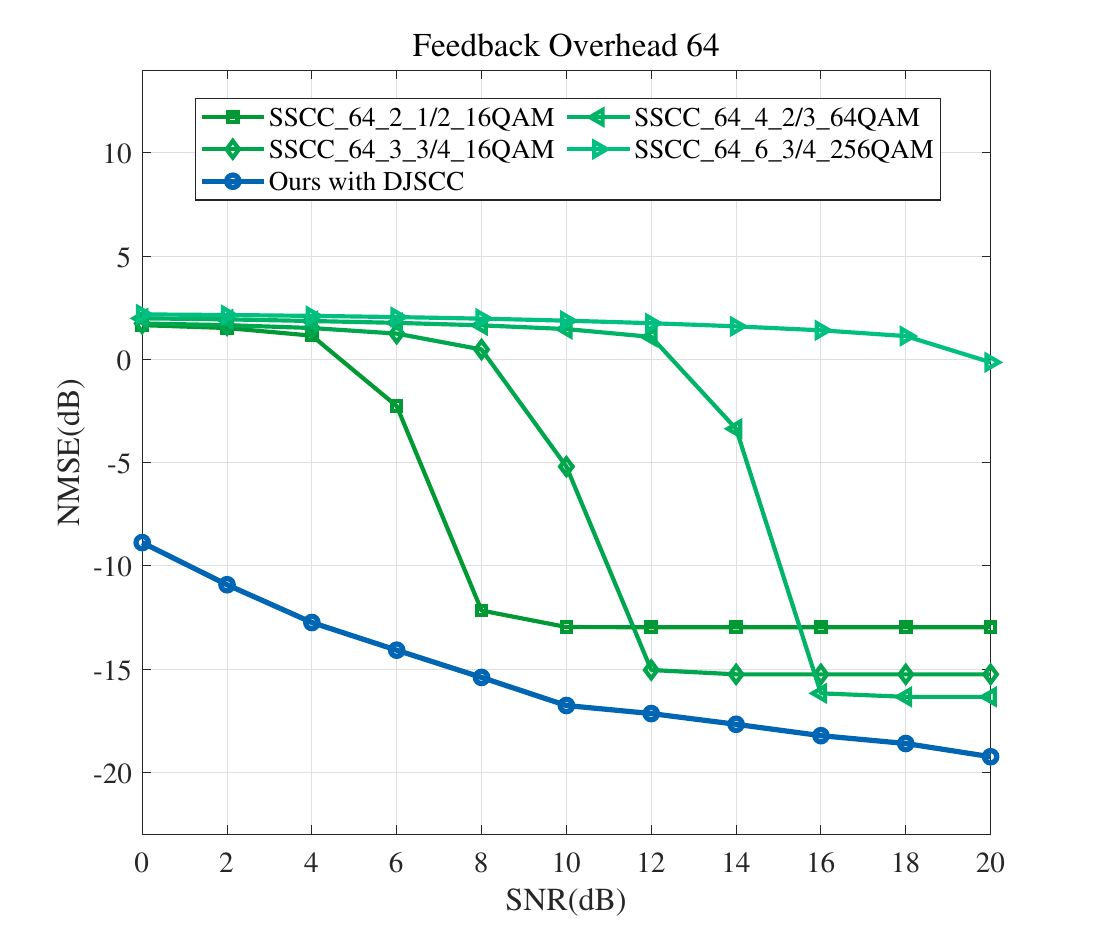}
		\vspace{-20pt} 
		\begin{center}
			\small{(a)} 
		\end{center}
		\label{fig:img1}
	\end{minipage}
	\begin{minipage}[b]{0.47\columnwidth}
		\centering
		\includegraphics[width=\textwidth]{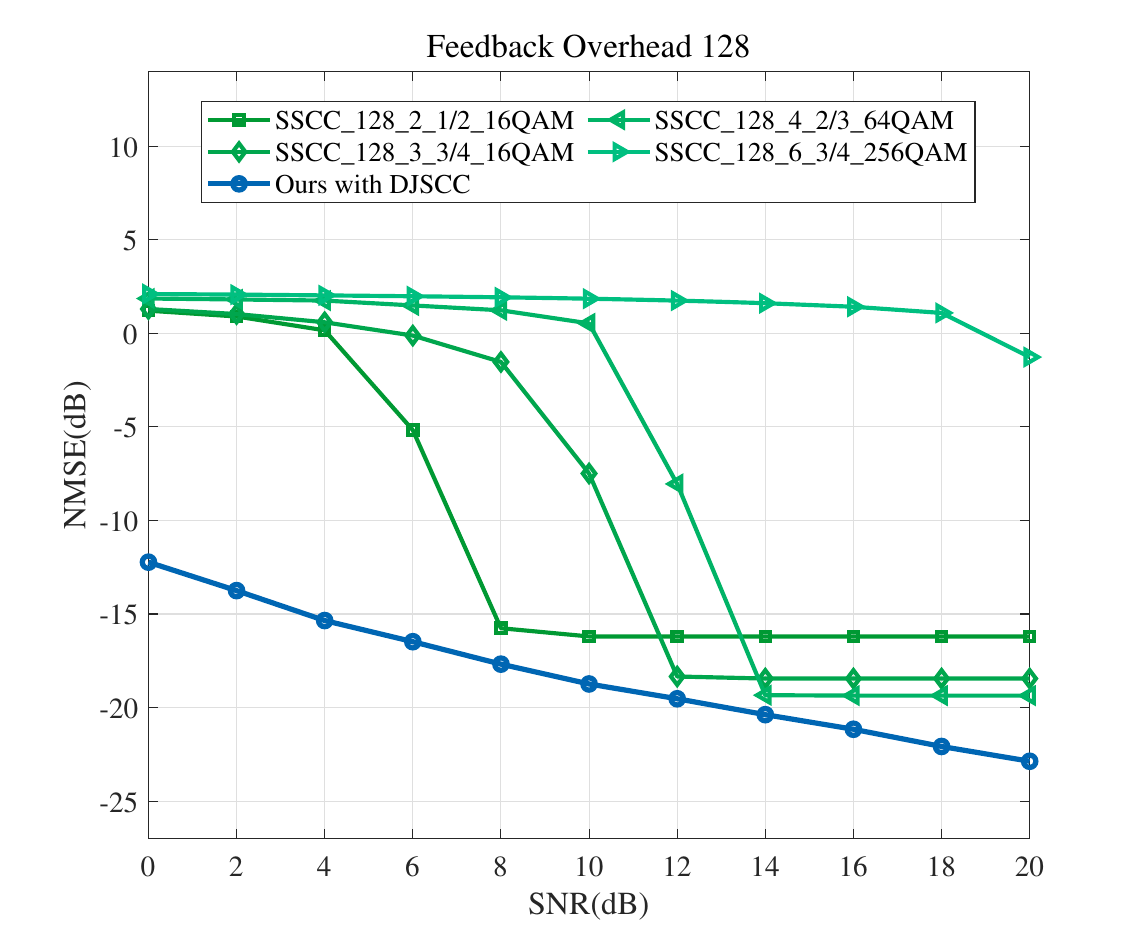}
		\vspace{-20pt} 
		\begin{center}
			\small{(b)} 
		\end{center}
		\label{fig:img2}
	\end{minipage}
	\vspace{-15pt}
	\caption{Performance comparison of the proposed multi-user network with DJSCC-based CSI feedback and conventional SSCC schemes under the 2-UE scenario: (a) feedback overhead 64; (b) feedback overhead 128.}
	\label{fig:SSCCvsJSCC}
	\vspace{-15pt}
\end{figure}
\subsection{Comparison with SSCC Feedback Scheme}
Fig. \ref{fig:SSCCvsJSCC} compare the reconstruction NMSE of the proposed DJSCC multi-user CSI feedback framework with conventional SSCC CSI feedback scheme under the 2-UE case. For SSCC, the same encoders and decoders as the proposed DJSCC is utilized. While the floating-point outputs of encoders are directly quantized to bit streams by a uniform quantization \cite{wang2022transformer} and fed back to the BS, which is then dequantized and processed by decoders. At the training stage, following \cite{xu2022deep}, a perfect bit-level feedback is assumed. At the testing stage, we employ low-density parity-check (LDPC) coding, and quadratic amplitude modulation (QAM) to transmit the feedback bits over the AWGN channel. The performance is evaluated under the same overhead. Specifically, our DJSCC's overhead is identical to the number of feedback symbols $d_k$, while the SSCC's overhead is ${d_k} = \left( {b \times q} \right)/\left( {r \times {{\log }_2}a} \right)$ symbols, where $b$ is the output dimension of the encoder in SSCC scheme, $q$ is the quantization bits, $r$ is the LDPC code rate and $a$ is the number of QAM constellation points.

We evaluated the methods with feedback overhead of 64 and 128 symbols at various SNRs $ \in \left[ {0,20} \right]$ dB. The SSCC methods are labeled as “SSCC\verb|_|b\verb|_|q\verb|_|r\verb|_|aQAM”. Results are shown in Fig.\ref{fig:SSCCvsJSCC}. As detailed in Section III-C, the DJSCC network was trained at random SNRs $ \in \left[ {0,20} \right]$ dB for 2000 epochs, followed by fine-tuning at fixed SNRs for fewer epochs (e.g., 200-400). This training approach slightly degrades NMSE performance but significantly reduces the training time. We can observe that SSCC methods exhibit “cliff-effect”: performance degrades sharply when the SNR falls below a certain threshold and no longer improves beyond another threshold, consistent with the effects observed in \cite{xu2022deep}. In contrast, the performance of DJSCC degrades slower as the SNR decreases and consistently outperforms the SSCC method across all SNRs, demonstrating the superiority of our DJSCC scheme.

\section{Conclusion}
In this letter, we proposed a multi-user CSI feedback framework called RCA-MUNet, whereby a residual cross-attention-based transformer is conceived to improve the CSI feedback in multi-user scenarios. Specifically, the proposed RCA-Block effectively exploits the CSI correlation among nearby UEs at the receiver side, considerably reducing the feedback overhead. We also extended the DJSCC scheme to multi-user cases and deployed a two-stage training strategy to improve the CSI feedback performance under varying conditions. Experimental results demonstrate that the RCA-MUNet not only achieves better performance, but also shows superiority in neural network complexity and scalability over state-of-the-art methods. Moreover, we highlight the effectiveness of DJSCC over conventional SSCC in multi-user feedback systems.

\bibliography{reference}

\end{document}